\documentclass{article}
\usepackage[dvipsnames, table]{xcolor}
\usepackage{hyperref}
\usepackage{epsfig, graphics, graphicx}
\usepackage{latexsym}
\usepackage{hyperref}
\usepackage{amsmath,amssymb,enumerate,comment}
\usepackage{graphicx}
\usepackage{caption}
\usepackage{subcaption}
\usepackage{color}
\usepackage[square,numbers]{natbib}
\usepackage{graphicx}
\usepackage{algorithm,algpseudocode}
\usepackage{times}
\usepackage{soul}
\usepackage{url}
\usepackage[margin=1in]{geometry}

\newcommand{\A}{{\boldsymbol{A}}}
\newcommand{\x}{{\boldsymbol{x}}}
\newcommand{\w}{{\boldsymbol{w}}}
\newcommand{\vv}{{\boldsymbol{v}}}
\newcommand{\ee}{{\boldsymbol{e}}}

\newcommand{\rr}{{\boldsymbol{r}}}
\newcommand{\y}{{\boldsymbol{y}}}
\newcommand{\supp}{\text{supp}}

\title{Data-driven Algorithm Selection and Parameter Tuning: Two Case studies in Optimization and Signal Processing}

\author{
Jes\'us A. De Loera \and Jamie~Haddock \and Anna~Ma \and Deanna~Needell}

\begin{document}

\maketitle

\begin{abstract}
Machine learning algorithms typically rely on optimization subroutines and are well-known to provide very effective outcomes for many types of problems. Here, we flip the reliance and ask the reverse question: can machine learning algorithms lead to more effective outcomes for optimization problems?
Our goal is to train machine learning methods to automatically improve the performance of optimization and signal processing algorithms. 
As a proof of concept, we use our approach to improve two popular data processing subroutines in data science: stochastic gradient descent and greedy methods in compressed sensing. 
 We provide experimental results that demonstrate the answer is ``yes'', machine learning algorithms do lead to more effective outcomes for optimization problems, and show the future potential for this research direction.
\end{abstract}

\section{Introduction} 

Machine learning is a popular and powerful tool that has emerged at the forefront of a vast array of applications (most famously 
in image processing).
At their core, neural nets rely on solving non-linear optimization problems.
From this point of view, improving key optimization subroutines and other auxiliary data processing methods directly 
helps to improve learning methods. 
Here, we aim to use machine learning algorithms to improve two optimization and signal processing subroutines,
which involve choice of parameters and algorithm selection.  
We believe our approach has great potential because these subroutines are central to the performance of
machine learning algorithms. Thus our set up leads to the intriguing ``meta" notion of using machine learning to improve machine learning. Our framework will be useful in other settings as well, where one must choose methods and/or parameters with limited knowledge about 
the input data.

In optimization and signal processing there are often choices among several algorithms or parameters to fine-tune in order to apply to an input instance.  
These choices can often lead to drastically different outcomes and thus such selections are crucial in many applications.
The questions we consider here are, \emph{what is the best way to select such algorithms?  What is the best choice of parameters?} 
In this paper, rather than using a one-size-fits-all rule to choose, we focus instead on the features of \emph{individual problem instances} 
and allow these features to guide the parameter or algorithm selection. 
 It is well-known that the performance of algorithms, even of those 
considered to be very efficient, depend on the particular input instances and data. Thus it makes sense to vary or adapt the 
choice of algorithm or parameters to the concrete instance in question, avoiding a one-size-fits-all approach. It is therefore natural
to use machine learning tools to perform the selection, just as a human expert could be making such choices.
 
Our work fits within a topic of Artificial Intelligence that has received several names: \emph{algorithm selection}, \emph{algorithm configuration}, \emph{self-adapting algorithms}, or simply \emph{automated machine learning (autoML)}.
This topic seeks to efficiently automate the selection of algorithms or their parameter configurations. It has attracted increasing attention, and relies on multiple techniques.
 Recent work on algorithm selection using machine learning has seen a strong surge in both practical and theoretical results.  
 Let us mention that \cite{lagoudakis+littman} approach the problem as a Markov Decision problem. In \cite{yang2018oboe}
the problem is approached with techniques similar to the matrix completion method. A learning framework for algorithm selection 
was presented in \cite{gupta2017pac} with a follow up in \cite{balcan}. From the computational and experimental point of view, 
 algorithm selection via machine learning has been used in several areas of optimization (see e.g., \cite{khaliletal2017, balcanetal18, andrychowicz2016learning} 
 and the many references therein). In fact, recently for the purpose of training algorithm selection, some libraries have been established to
organize data for a wide range of NP-hard tasks (where the aim is to predict how long an algorithm will take to solve concrete instances of NP-complete problems, or to choose best approximation schemes tailored by instances) \cite{nudelman04,Bischletal16,kotthoffHO17}. While these works are combinatorial in nature, here we propose that learning can drive optimal selection in continuous and analytical problems.

\subsection*{\textbf{Our contributions}}

In this work we study the problems of algorithm selection 
in \emph{compressed sensing} and parameter tuning for SGD algorithms. In both cases we wish to train a recommendation or classification 
algorithm that can output an optimal algorithm (method, parameters, etc.) for a given input data set.
Note that in contrast to previous work in step size tuning for SGD, our work does not require that iteration specific step sizes be computed at every step.
We study these problems from the experimental point of view. Our main contributions are as follows.

\begin{itemize} 

\item In Section \ref{section-cs} we apply our methodology, through concrete experiments, to the selection of compressed sensing algorithms.
Here we concentrate on selecting the best among three well-known greedy algorithms for solving the compressed sensing problem: \emph{Hard 
Thresholding Pursuit} (HTP)  \cite{foucart2011hard}, \emph{Normalized Iterative Hard Thresholding} (NIHT) \cite{blumensath2010normalized}, and \emph{Compressive Sampling Matching Pursuit with Subspace 
Pursuit} (CSMPSP) \cite{NeedeT_CoSaMP,maleki2010optimally}. We have been inspired by the work of \cite{blanchard2015performance}, where the authors catalog optimal algorithm selection through brute force experimental testing. Although our machine learning approach is useful precisely when such a rigorous catalog is \textit{not} available, the work of \cite{blanchard2015performance} will be used as validation of our framework. 
 
\item In Section \ref{section-sgd} we apply our methodology, through concrete experiments, to the selection of the best step size in the popular \emph{stochastic gradient descent algorithm} \cite{sgd}, which itself is used as a subroutine in many learning frameworks. Unfortunately, tuning the step size (also called the \textit{learning rate}) is often more an art than a science, and the selection can lead to drastically different overall behavior. We aim to alleviate this issue by allowing for such selections to be done by the trained machine.

\end{itemize}

For our experiments we use Neural Networks for classification. Neural Networks are computing systems inspired by the biological neural networks. They have shown remarkable success in various machine learning tasks including classification \cite{krizhevsky2012imagenet}. While there are plenty of sophisticated, state of the art neural net architectures such as GoogLeNet \cite{szegedy2015going}, 
ResNet \cite{he2016deep}, DenseNet \cite{huang2017densely} and CliqueNet \cite{yang2018convolutional}, we will demonstrate that even simple networks that do not have to be run on expensive remote processors can aid in algorithm selection. 
This neural net learning approach is perfect for learning and modeling non-linear and complex relationships allowing, as the name suggests, the machine to learn data relationships by itself.  

While there has been much work in the area of (autoML) and automated algorithm selection, these automated approaches differ from ours in simplicity of application to new data. These methods require a preprocessing step before application of the learning technique for algorithm selection \cite{balte2014meta,bardenet2013collaborative, feurer2015efficient,pfahringer2000meta,yang2018oboe}.  Meanwhile, our approach simply uses the data encoding the problem (or even simpler attributes of the data) as input features to our learning approach. This straight-forward approach allows practitioners to apply this algorithm selection framework without the expertise to determine meta-features of the data that will enable an effective learning approach.

\subsection*{Notation}
Here and throughout the paper, we write
\begin{equation}
\A\x = \y,
\label{eq:linearsystem}
\end{equation}
where $\A \in \mathbb{R}^{m \times n}$ is the measurement (or data) matrix, $\y \in \mathbb{R}^{m}$ is the measurement vector, 
and $\x\in\mathbb{R}^n$ is the signal being recovered. We use $(\cdot)^*$ to denote the transpose operator and $(\cdot)^\dagger$ 
denotes the pseudo-inverse.

In the compressed sensing problem considered in Section \ref{section-cs}, the measurement matrix is underdetermined ($m \ll n$) and the signal is assumed to be sparse; in particular, we say $\x$ is $s$-sparse when it has at most $s$ non-zero entries. Furthermore, for any vector $\vv$, $\supp_s(\vv)$ returns the indices corresponding to the $s$ largest in magnitude entries of $\vv$ and $P_T(\vv)$ returns a vector whose entries are 0 outside of the support of set $T$ and equal to $\vv$ on the support of $T$. For any set $T\subset \{1, \cdots, n \}$, a matrix $\A$ constrained to the columns indexed by $T$ is denoted by $\A_{T} \in \mathbb{R}^{m \times |T|}$. 

In the least-squares problems considered in Section \ref{section-sgd}, the measurement matrix is overdetermined ($m \gg n$) and no sparsity assumption is made on the signal $\x$. We use the recovery error and the residual error at the $M$-th iteration of SGD, $\|\x_M - \x\|_2$ and $\|\A\x_M - \y\|_2$ respectively, to measure the performance of the algorithm with given learning rates.

In both Sections \ref{section-cs} and \ref{section-sgd}, we will make use of a maximum iteration cap to stop the algorithm in question, which we denote $M$. The use of this stopping criterion is described in detail in each section.

\section{Application I: compressed sensing algorithm selection} 
\label{section-cs}

We begin the investigation of our framework with a proof of concept inspired by the work done in~\cite{blanchard2015performance} where the authors rigorously test various compressed sensing methods under various settings in a brute-force way. We will show that we can use neural networks to recover the phase transitions that were acquired via rigorous testing in the aforementioned paper. For this reason, we adopt a similar algorithmic and experimental setup. 
First, we will explain the compressed sensing problem and notation used throughout, then we will present three greedy algorithms. Following that, the experimental setup including the different sensing matrices, signal initialization, and stopping criteria are discussed. Finally, we present our experimental results and remark on our findings.

There is now an abundance of both theory and algorithms that guarantee robust and accurate recovery of sparse signals, under various assumptions on the measurement matrix $\A$ \cite{foucart2013mathematical,eldar2012compressed}. For example, the so-called \textit{Restricted Isometry Property} \cite{RefWorks:48} guarantees such recovery and random matrix constructions are shown to satisfy this property when the number of measurements $m$ scales like $s\log n$ \cite{RefWorks:285}. Under this or related assumptions, both greedy (iterative) algorithms and optimization-based methods (e.g., L1-minimization) are shown to produce accurate recovery results. In general, the performance of such algorithms depend on the undersampling and oversampling rates which we denote as 
\begin{equation}
\delta = \frac{m}{n} \text{ and } \rho  = \frac{s}{m},
\label{eq:deltarho}
\end{equation}
respectively. Furthermore, we refer to combinations of $\delta$ and $\rho$ as the \emph{($\delta, \rho$) plane}. By observing the behavior of algorithms on the $(\delta, \rho)$ plane, we can see how different approaches act under various sampling rates.

We consider three greedy algorithms for solving the compressed sensing problem: \emph{Hard Thresholding Pursuit} (HTP) \cite{foucart2011hard}, \emph{Normalized Iterative Hard Thresholding} (NIHT) \cite{blumensath2010normalized}, and \emph{Compressive Sampling Matching Pursuit with Subspace Pursuit} (CSMPSP) \cite{NeedeT_CoSaMP,maleki2010optimally}. The pseudo-code for HTP, NIHT, and CSMPSP appears in Algorithm~\ref{alg:htp}, Algorithm~\ref{alg:niht}, and Algorithm~\ref{alg:CSMPSP} respectively. These methods are all similar in spirit; they seek to recover the signal $\x$ from $\y$ while also identifying the support of $\x$, which is discovered iteratively. Each essentially uses a \textit{proxy} for the signal $\x$ (e.g., $\A^*\y$) to identify a support estimate $T$, then estimates $\x$ on that support (e.g., $\x_t=\A_T^\dagger \y$) , then computes the residual and repeats the process to locate the remainder of $\x$. HTP and NIHT use specially chosen step sizes (denoted $w_k$) when updating the estimate to $\x$ and recompute the support in each iteration, whereas CSMPSP uses a union of prior estimates followed by pruning. See Algorithms \ref{alg:htp}-\ref{alg:CSMPSP} and \cite{foucart2011hard,blumensath2010normalized,maleki2010optimally} for details about these approaches. What is important for our purpose is that each algorithm may perform differently for a given set of inputs, leading to varying accuracy on the output. Therefore, there is value in using machine learning tools to decide what is the best choice of algorithm in a given problem instance. Also note that each algorithm takes the same inputs, namely the measurement matrix $\A$, the measurement vector $\y$, and an approximation for the number of nonzero entries $s$ in the sparse signal.

Although the theory for these approaches holds \textit{uniformly}, meaning it holds for any sparse signal and matrix satisfying the assumptions, it has long been observed that the algorithms actually behave quite differently on various kinds of signal and measurement ensembles \cite{practicalGGKLNT14,Paper5,blanchard2015performance}. In fact, \cite{blanchard2015performance} documents an extensive comparison of these approaches for various ensembles while ranging the parameters $\delta$ and $\rho$. This latter work can be used as a ``lookup table," when one knows the input information and wants to select the optimal algorithm for their purpose. Their work, in some sense, motivates us to apply the machine learning methodology to compressed sensing, as we have a comprehensive benchmark with which to compare these methods. Note that these comparisons were made in a brute force manner, where each method was run on each ensemble type over a fine grid of input parameters. Such an exhaustive approach is not practical when the input domain is extremely large. Moreover, in this setting, we have a greater understanding of how these greedy algorithms will behave for a specific problem instances, making it an appropriate problem to verify and validate our framework.

\begin{algorithm}[t!]
	\caption{Hard Thresholding Pursuit}\label{alg:htp}
	\begin{algorithmic}[1]
		\Procedure{HTP}{$\A, \y, s$}
		\State Initialize $\x_0 = \A^*\y$, $T_0 = \supp_s(\x_0)$, {$\x_0 = P_{T_0}(\x_0)$}, $\rr_0 = \y - \A\x_0$, $k=1$
		\While{stopping criteria not reached}
		\State $w_k = \frac{\| (\A^*\rr_{k-1})_{T_{k-1}}\|_2^2}{\|\A_{T_{k-1}}(\A^* \rr_{k-1})_{T_{k-1}}\|_2^2}$
		\State $\x_k = \x_{k-1} + \w_k\A^*\rr_{k-1} $
		\State $T_k = \supp_s(\x_k)$
		\State $\x_k = \A_{T_k}^\dagger$y
		\State $\rr_k = \y - \A\x_{k-1}$
		\State $k=k+1$
		\EndWhile
		\EndProcedure
	\end{algorithmic}
\end{algorithm}

\begin{algorithm}[t!]
	\caption{Normalized Iterative Hard Thresholding}\label{alg:niht}
	\begin{algorithmic}[1]
		\Procedure{NIHT}{$\A, \y, s$}
		\State Initialize $\x_0 = \A^*\y$, $T_0 = \supp_s(\x_0)$, {$\x_0 = P_{T_0}(\x_0)$}, $\rr_0 = \y - \A\x_0$, $k=1$
		\While{stopping criteria not reached}
		\State $w_k = \frac{\| (\A^* \rr_{k-1})_{T_{k-1}}\|_2^2}{\|\A_{T_{k-1}}(\A^* \rr_{k-1})_{T_{k-1}}\|_2^2}$
		\State $\x_k = \x_{k-1} + \w_k\A^*\rr_{k-1} $
		\State $T_k = \supp_s(\x_k)$
		\State $\x_k = P_{T_k}(\x_k)$
		\State $\rr_k = \y - \A\x_{k}$
		\State $k=k+1$
		\EndWhile
		\EndProcedure
	\end{algorithmic}
\end{algorithm}

\begin{algorithm}[t!]
	\caption{Compressive Sampling Matching Pursuit with Subspace Pursuit}\label{alg:CSMPSP}
	\begin{algorithmic}[1]
		\Procedure{CSMPSP}{$\A ,\y, s$}
		\State Initialize $\x_0 = \A^*\y$, $T_0 = \supp_k(\x_0)$, $\x_0 = P_{T_0}(\x_0)$, $\rr_0 = \y - \A\x_0$, $k=1$
		\While{stopping criteria not reached}
		\State $S_k = \supp(\A^*\rr_{k-1})$
		\State $\Lambda_k = T_{k-1} \cup S_k$
		\State $\x_k = \A_{\Lambda_k}^\dagger\y$
		\State $T_k =\supp_s(\x_k)$
		\State $\x_k = P_{T_k}(\x_k)$
		\State $\rr_k = \y - \A\x_k$
		\State $ k = k+1$
		\EndWhile
		\EndProcedure
	\end{algorithmic}
\end{algorithm}

\subsection{Experimental setup}
\label{sec:cs-exp-setup}
We consider three randomly generated measurement matrices for this setting: Gaussian, Sparse, and Discrete Cosine Transform (DCT). Entries of the Gaussian matrices are drawn $i.i.d.$ from $\mathcal{N}(0, \frac{1}{m})$ so that in expectation, they have normalized columns. Sparse measurement matrices have $p=7$ nonzero entries in each column where the value of the nonzero entries is drawn from $\{\pm p^{-\frac{1}{2}} \}$ with equal probability. Finally the DCT measurement matrices consist of $m$ randomly subsampled rows of the $n \times n$ full DCT matrix. The number of measurements $m$ is determined by $\delta$ and the vector $\x$ being recovered has $s$ nonzero entries (determined by $\rho$) and takes on values $\{\pm 1\}$ with equal probability where $\delta$ and $\rho$ are as defined in \eqref{eq:deltarho}. The measurement vector $\y = \A\x$ where $\A$ is one of the three types of measurement matrices and $\x$ is the signal to be recovered.

We terminate any algorithm when it satisfies one of the following stopping criteria. 
\begin{enumerate}
\item \textbf{Convergence} - An algorithm is convergent if the residual error is small enough. In particular, if $$ \| \A \x_k - \y \| \leq 0.001 \delta. $$
\item \textbf{Divergence} - An algorithm is divergent if the residual error is larger than a factor of the norm of the initial residual: 
$$  \| \A \x_k - \y \| \geq 100  \| \A \x_0 - \y \|. $$
\item \textbf{Slow Progress I} - After $750$ iterations of NIHT or $150$ iterations of CSMPSP or HTP, we begin to check for slow progress. For the first version of ``slow progress" we check whether the residual has made any significant progress over the last 15 iterations:
$$ \max_i | \| \A \x_{k-i+1} - \y \| - \| \A \x_{k-i} - \y \| | \leq 10^{-6}. $$
\item \textbf{Slow Progress II} - After $750$ iterations of NIHT or $150$ iterations of CSMPSP or HTP, we check whether the convergence rate is close to 1: 
$$ \left( \frac{\|\A \x_{k-15} - \y \| }{\| \A \x_k -\y \|} \right) ^\frac{1}{15} \leq 0.999.$$ 
\item \textbf{Maximum Iteration} - An algorithm that runs for longer than 60 minutes (discounting time for computing metrics) or $M$ iterations (where $M = 900$ for NIHT and $M = 200$ for CSMPSP and HTP) has reached the allowable computation time and is terminated.
\end{enumerate}
It should be noted the algorithm stopping criteria of (1)-(4) are as in~\cite{blanchard2015performance} while the last exit was added to keep from a single experiment from running for too long. Practically, the last stopping criteria reflects a computational time constraint.

\subsection{Experiments}
In the following set of experiments, we train neural networks to classify whether or not an algorithm can recover a signal in the standard compressed sensing problem \eqref{eq:linearsystem}. The experiment requires three phases: creating training data, training the neural network, and testing the neural network.

In the first phase, training data with labels are created to input into the neural network. The training data set comprises of 2241 samples. For each matrix type (Gaussian, Sparse, DCT), there are 747 training points on the $(\delta,\rho)$ plane (See \eqref{eq:deltarho}). For each $(\delta,\rho)$ pair, we run Algorithm~\ref{alg:htp}, Algorithm~\ref{alg:niht}, and Algorithm~\ref{alg:CSMPSP} until the algorithm satisfies one of the stopping criteria discussed in Section~\ref{sec:cs-exp-setup}. In order for a given algorithm to be labeled as ``successful" at recovering signals for a specified $(\delta,\rho)$ and measurement matrix, 50 of the 100 randomly generated samples must have satisfied the ``convergence" stopping criteria. This phase is completed in MATLAB using version R2014b on a desktop running Linux.
 
The training data from the first phase and labels are used to train neural networks in the second phase. The input variables used by the neural network are the signal dimension $n$, the number of measurements $m$, the number of non-zero entries $s$, and an indicator variable that indicates the measurement matrix. The second phase is accomplished using Python 3 and Keras 2.1 with TensorFlow as a backend. We set up the neural network to contain two hidden layers, the first with three nodes and a second layer with nine nodes, and offer the following intuition for the neural network structure. The purpose of the first layer is to determine the measurement matrix type while the second layer classifies whether or not an algorithm will be successful. The hidden layers utilize ReLu as their activation function, with the exception of the final output layer which uses the sigmoid function. Approximately 90\% of the available data is used to train our neural network for each algorithm and the remaining 10\% is used to measure validation accuracy on the trained network.

\begin{figure}[ht!]
\centering
\includegraphics[width=.7\textwidth]{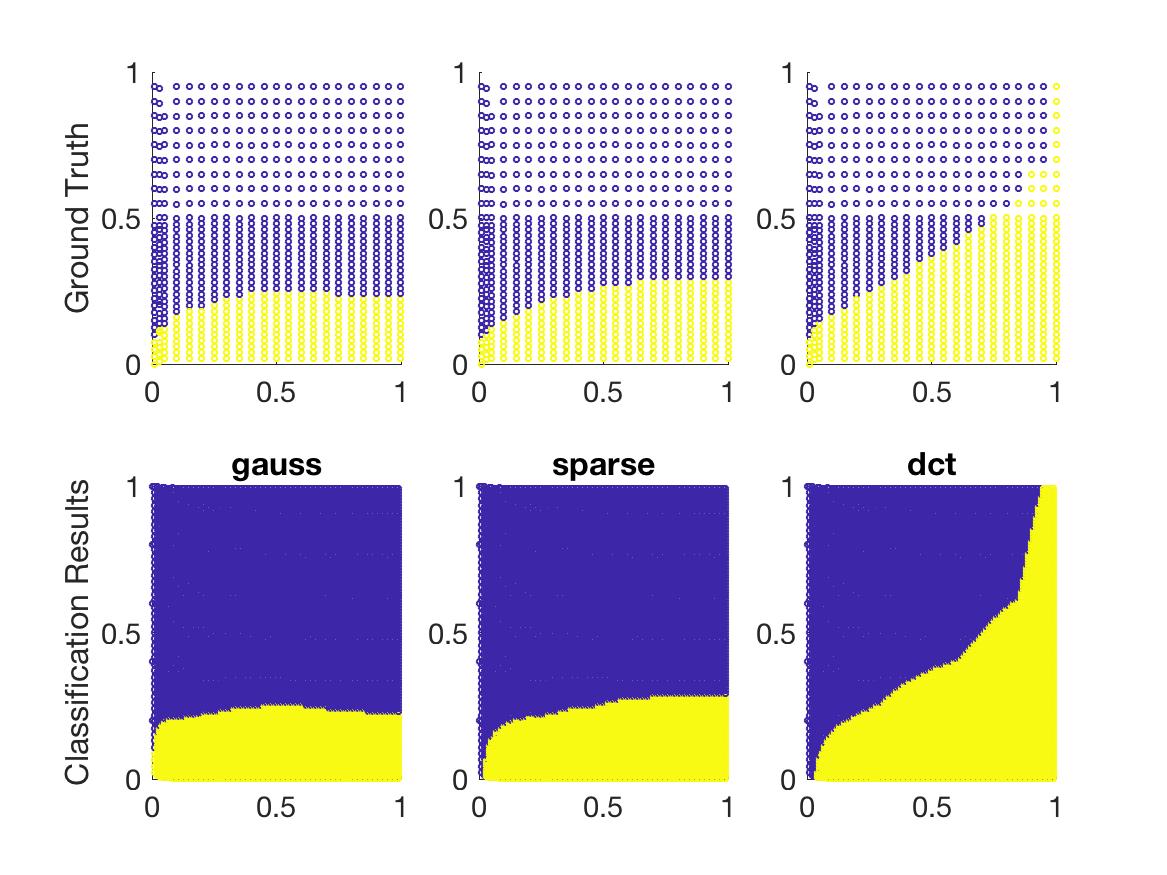}
\caption{Recovery for phase transitions of various types of measurement matrices for HTP.  (Test validation accuracy = 0.969)}
\label{fig:htp}
\end{figure}

\begin{figure}[ht!]
\centering
\includegraphics[width=.7\textwidth]{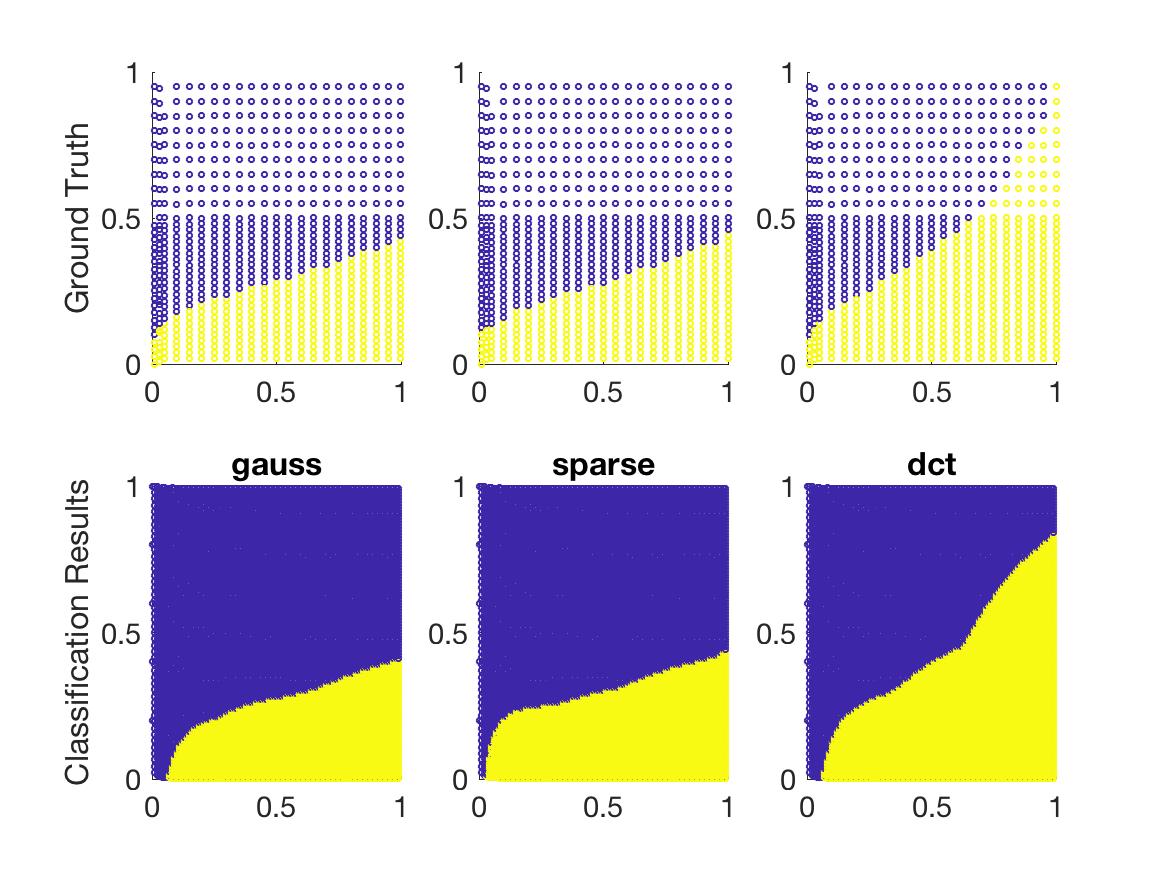}
\caption{Recovery for phase transitions of various types of measurement matrices for NIHT. (Test validation accuracy = 0.964)}
\label{fig:niht}
\end{figure}

\begin{figure}[ht!]
\centering
\includegraphics[width=.7\textwidth]{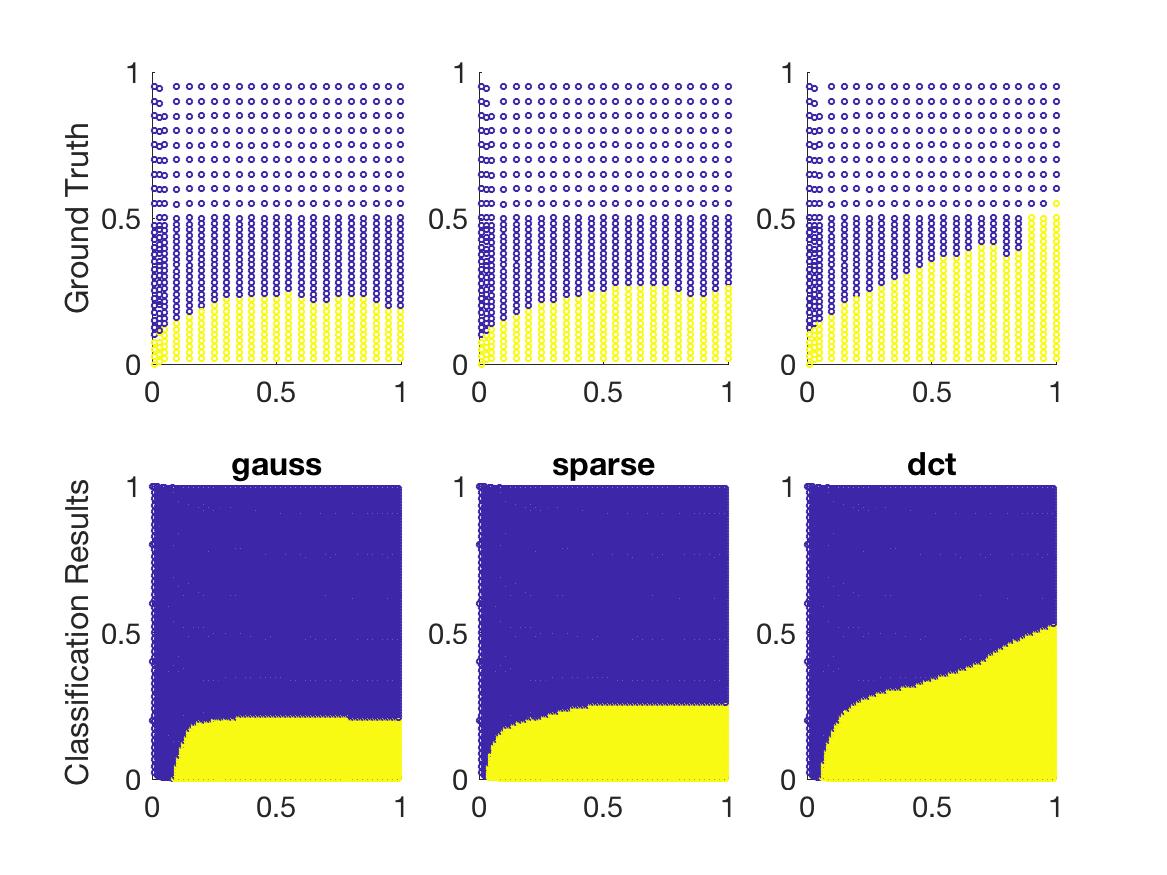}
\caption{Recovery for phase transitions of various types of measurement matrices for CSMPSP. (Test validation accuracy = 0.963)}
\label{fig:csmpsp}
\end{figure}

Figures~\ref{fig:htp}, \ref{fig:niht}, and \ref{fig:csmpsp} present the computational results for HTP, NIHT, and CSMPSP respectively. In each subplot, the horizontal axis represents the value of $\delta$ and the vertical axis represents the value of $\rho$. Furthermore, each figure can be broken down as follows. Each column isolates a specific measurement matrix: Gaussian, sparse, and DCT (left to right). The first row of each figure shows the training $(\delta,\rho)$ pairs (i.e., data created in first phase of experiment) along with their labels, indicated by the color of the data point. Here, yellow points indicate that an algorithm is ``successful" and blue points indicate that the algorithm is not successful. In the second row of each figure, we show results produced by the trained neural network from the second phase on test data created by uniformly sampling the $(\delta,\rho)$ plane. The accuracy of the trained networks on validation data is reported in the captions of each figure. For all experiments, the signal dimension $n = 2^{12}$ while $m$ and $k$ are computed according to the specified $\delta$ and $\rho$. 

These numerical experiments show that even a simple neural network is able to approximately determine whether or not a given greedy algorithm and $(\delta, \rho)$ pairing will result in successful signal recovery. In particular, the yellow regions in the second row of each figure not only roughly approximate the yellow regions in the first row but they also noticeably vary across both algorithm and measurement matrix to match the input training data, as desired. 

\section{Application II: stochastic gradient descent learning rate selection}\label{section-sgd}

We now further test our machine learning framework with an exploration of learning rate schedule selection for the \emph{stochastic gradient descent} (SGD) algorithm.  In this set of experiments, we demonstrate that one can use neural networks to select a learning rate schedule which improves the behavior of SGD on a given instance, provided proper training data.  After a brief introduction to the vast body of literature regarding the convergence behavior of SGD and corresponding learning rates, we discuss our experimental results and comment on our findings.  In Subsection \ref{subsec:sgd-experiments}, we describe in detail the design of our neural network framework.  We additionally describe the construction of the training and testing data provided to the network in each experiment.

SGD is an ubiquitous first-order iterative method for convex optimization.  The classical SGD algorithm for optimizing $f(x)$ works as follows: After selecting a learning rate (or step size) schedule $\alpha_t$ and an initialization $\x = \x_0$, we randomly select an index $i ~ \in \{1, ..., m \}$. While the stopping criteria is not satisfied we update $\x_t = \x_{t-1} - \alpha_t \nabla f_i(\x_{t-1})$. The applications in which SGD is \emph{la m{\'e}thode du jour} are diverse and cut across many scientific fields, with perhaps the hottest application currently being in the training of neural networks.  The performance of SGD depends heavily on the selected learning rate (or step size) schedule, $\{\alpha_t\}_{t = 1}^M$, and parameters of the objective function such as the Lipchitz constant or strong-convexity parameter \cite{sgd, shamir2013stochastic, moulines2011non, needell2014stochastic}. Parameter tuning SGD can also be interpreted as an algorithm selection problem. There are numerous proposed line search methods for selecting learning rates and methods for performing one-dimensional optimization on the learning rate to speed convergence \cite{masse2015speed,de2016big,mahsereci2015probabilistic,tan2016barzilai}. In practice, learning rate selection can be quite ad-hoc and there are popular heuristics for updating the learning rate \cite{goyal2017accurate}.

Recently, practitioners and theorists alike have turned their attention to adaptive learning rate schedules, in which the learning rate assigned to a component updates according to information gleaned from the sample \cite{duchi2011adaptive, zeiler2012adadelta, schaul2013no, kingma2014adam, defossez2017adabatch}. Recent adaptive learning rate approaches approximate Lipschitz parameters and use this to approximately compute a learning rate \cite{needell2014stochastic, wu2018wngrad}.  

Our work presents a machine learning framework which allows practitioners to choose a learning rate schedule without knowledge of objective function parameters.  As a proof of concept, we focus on solving least-squares problems, but we stress that our framework could be applied to more complex objective functions. This framework offers practitioners an alternative to heuristics and unknown objective function parameters.

\subsection{Experiments}
\label{subsec:sgd-experiments}
In each of the experiments presented below, we apply SGD to solve a least-squares problem $\|\A\x - \y\|_2^2$ defined by measurement matrix $\A$ and measurement vector $\y$.  The goal of our machine learning framework is to train a neural network to select the optimal learning rate schedule (out of a fixed set of schedules) for a given input linear system represented by its measurement vector, $\y$; we specify the measure with which we compare learning rates in each section below. These experiments also require the same three phases as in Section \ref{section-cs}: creating training data, training the neural network, and testing the behavior of SGD with the neural network predicted learning rates.

In the first phase, we generate data points consisting of measurement vectors $\y$ and labels that indicate the optimal learning rate schedule. We compare only two types of learning rate schedules: the constant learning rate $\alpha_t = c$ and the epoch-based learning rate schedule $\alpha_t = c_1*c_2^{\lfloor t+1/c_3 \rfloor}$; these constants are defined in each experiment below.  To select the optimal learning rate schedule and assign a label to each data point, we run $M = 5000$ iterations of SGD and assign the label of the learning rate schedule that resulted in the smallest recovery error, $\|\x_M - \x\|_2$ where $\x$ is the signal.  This phase is completed in MATLAB using version R2017a on a laptop running macOS. In each experiment, our input data points form two classes which correspond to each of the learning rate schedules.  The consistent systems are optimally solved with the constant learning rate schedule, while the inconsistent systems are optimally solved with the epoch-based learning rate schedule; this decreasing learning rate schedule helps SGD avoid the larger convergence horizon of the inconsistent systems. These data points are labeled accordingly and the neural network task of predicting the optimal learning rate schedule is equivalent to predicting to which set of systems each data point belongs. 
In each experiment, the data set consists of 3000 measurement vectors, a portion of which is used for training and the remaining data set is reserved for testing.  

In the second phase, we train a neural network with the training portion of the data set, consisting of the measurement vectors $\y$ and the optimal learning rate schedule labels for each system.  The second phase is performed in Python 3 and Keras 2.1 with TensorFlow as a backend.  The neural network architecture we adopt has one hidden layer with $30$ nodes.  The intuition for this choice of network architecture is that in our experimental setup the network only needs to determine which systems are consistent; as a linear problem, we expect that a thin, simple architecture should be successful. The hidden layer nodes use ReLU as the activation function and the final output layer uses the sigmoid function.  In the experiments below, we sample $75\%, 50\%$, and $25\%$ of the data to train the neural network and reserve the remaining data for testing validation accuracy.   

In the third phase, we measure the validation accuracy of the trained neural network predictions on the test set.  Additionally, we use the neural network predicted learning rate schedules to solve each least-squares problem in the test set with $M = 5000$ iterations of SGD and measure the resulting average recovery error, $\|\x_M - \x\|_2$, and average residual error, $\|\A\x_M-\y\|_2$ over the test set.  We compare these average error measures for the neural network predicted learning rates with the average errors solving the test set using only the constant learning rate schedule and only the epoch-based learning rate schedule.

\subsubsection{Synthetic Linear Systems}

In this experiment, we train a neural network to recommend either the constant learning rate $\alpha_t = 0.01$ or the epoch-based learning rate schedule $\alpha_t = 0.01 * 0.3^{\lfloor t+1/100\rfloor}$.  Here we set $\A$ to be a fixed $1000 \times 100$ matrix with Gaussian random variable entries drawn \emph{i.i.d.} from $\mathcal{N}(0,1)$, and we design two types of linear systems with this matrix, consistent and inconsistent.  For the set of consistent linear systems, we set $\y = \A\x/\|\A\x\|_2$ where $\x$ is a Gaussian vector.  For the set of inconsistent systems, we set the error $\ee = \vv - \A(\A^*\A)^{\dagger}\A^*\vv$ where $\vv$ is a Gaussian random variable, so that $\ee$ is orthogonal to the column space of $\A$.  We then set $\y = \A\x/\|\A\x\|_2 + \ee/\|\ee\|_2$ and normalize so that $\y$ has $\|\y\|_2 = 1$.  The set of consistent systems are optimally solved with the constant learning rate schedule and the set of inconsistent systems are optimally solved with the epoch-based learning rate schedule.  

We train the neural network with random subsets of a collection of 3000 linear system measurement vectors $\y$, 1500 of which are consistent and 1500 of which are inconsistent.  In our experiment, we measure the average validation accuracy of the neural network predictions on the remaining test measurement vectors for ten trials in which we randomly sample subsets of $75\%$, $50\%$, and $25\%$ training data of the 3000 measurement vectors; the average validation accuracies are listed in Table \ref{table:Gaussortho}. 
Furthermore, we list the average recovery error, $\|\x_M - \x\|_2$, and average residual error, $\|\A\x_M-\y\|_2$, for the approximation computed by $M = 5000$ SGD iterations using first the constant learning rate schedule, then the epoch-based learning rate schedule, and finally the neural network predicted learning rate for each system.  These measures are listed in Table \ref{table:Gaussortho}; the smallest average error is bolded in each row. Note that the average recovery error and average residual error for the neural network predicted learning rates are lower than those of the constant learning rate or epoch-based learning rate for the neural networks trained with $75\%$ and $50\%$ of the data.  We suspect that the errors associated with the learning rates predicted by the neural network trained with $25\%$ of the data are not the lowest because of the low neural network validation accuracy, which is in turn due to the small amount of training data.  

\begin{table}[t!]
	\begin{center}
		\begin{tabular}{c || c | c | c | c }
			\hline
			Train & Validation & \multicolumn{3}{c}{$\|\x_M - \x\|_2$} \\
			$\%$ & Accuracy & Const. & Epoch & NN Pred.\\
			\hline
			$75\%$ & 86.00\% & 0.01142 & 0.01525  & \textbf{0.00909} \\
			$50\%$ & 77.01\% & 0.01138 & 0.01530 & \textbf{0.01064} \\
			$25\%$ & 66.32\% & \textbf{0.01116} & 0.01524 & 0.01177 \\
			\hline
		\end{tabular}
		\begin{tabular}{c || c | c | c }
			\hline
			Train &  \multicolumn{3}{c}{$\|\A\x_M - \y\|_2$}\\
			$\%$ & Const. & Epoch & NN Pred.\\
			\hline
			$75\%$ & 0.50980 & 0.64512 & \textbf{0.45912} \\
			$50\%$ & 0.50806 & 0.64590 & \textbf{0.49707} \\
			$25\%$ & \textbf{0.49739} & 0.64027 & 0.53053 \\
			\hline
		\end{tabular}
	\end{center}
	\caption{Average test set validation accuracies of trained neural network (averaged over 10 trials), average resulting recovery error $\|\x_M - \x\|_2$ and residual error $\|\A\x_M - \y\|_2$ on test set for constant learning rate, epoch-based learning rate, and the neural network (NN) predicted learning rates on synthetic linear systems.}
	\label{table:Gaussortho}
\end{table} 

\subsubsection{Computerized Tomography Systems} 

In this experiment, we again train a neural network to recommend either the constant learning rate $\alpha_t = 0.01$ or the epoch-based learning rate schedule $\alpha_t = 0.01*0.95^{\lfloor t+1/100 \rfloor}$.  We input two types of linear systems, consistent and inconsistent.  Each data point input is the measurement vector $\y$ from a computerized tomography system of equations, $\A\x = \y$ (generated by code adapted from the regularization toolbox by PC Hansen \cite{hansen1994regularization}).  We fix the matrix $\A$ to be a CT matrix generated by the command \texttt{tomo(20,10)}; here $N = 20$ is the discretization parameter (number of pixels along one edge of the square image) and $f=10$ is the oversampling factor.  This matrix represents the ray directions which are sampled through the signal (image).  We then produce consistent CT systems by applying the CT matrix $\A$ to the signal $\x$, which is an image from the MNIST database \cite{MNIST}, producing the measurement vector $\y = \A\x$ and then normalizing so that $\|\y\|_2 = 1$. These measurement vectors contain a linear combination of the pixels through which the tomography rays pass. This set of systems is optimally solved with the constant learning rate.  We produce inconsistent CT systems with error $\ee = \vv- \A(\A^*\A)^{\dagger}\A^*\vv$ where $\vv$ is a Gaussian random variable, so that $\ee$ is orthogonal to the column space of $\A$.  The measurement vector $\y$ for these inconsistent CT systems is $\y = \A\x/\|\A\x\|_2 + 0.5*\ee/\|\ee\|_2$ normalized so that $\|\y\|_2 = 1$, where $\x$ is an image from the MNIST database.  This set of systems is optimally solved with the epoch-based learning rate schedule.  

To evaluate these methods, we measure the average validation accuracy of the neural network predictions on the remaining test measurement vectors for ten trials in which we randomly sample subsets of $25\%$, $50\%$, and $75\%$ training data of the 3000 measurement vectors; the average validation accuracies are listed in Table \ref{table:CTortho}.  Furthermore, we list the average recovery error, $\|\x_M - \x\|_2$, and average residual error, $\|\A\x_M-\y\|_2$, for the approximation computed by $M = 5000$ SGD iterations using first the constant learning rate schedule, then the epoch-based learning rate schedule, and finally the neural network predicted learning rate for each system.  These measures are listed in Table \ref{table:CTortho}; the smallest average error is bolded in each row. Note that the average recovery error and average residual error for the neural network predicted learning rates are lower than those of the constant learning rate or epoch-based learning rate, except for the average residual error of the neural network trained with $50\%$ of the data. 

\begin{table}[t!]
	\begin{center}
	\begin{tabular}{c || c | c | c | c }
		\hline
		Train & Validation & \multicolumn{3}{c}{$\|\x_M - \x\|_2$} \\
		 $\%$ & Accuracy & Const. & Epoch & NN Pred.\\
		 \hline
		 $75\%$ & 88.19\% & 0.00669 & 0.00687 & \textbf{0.00584} \\
		 $50\%$ & 79.68\% & 0.00669 & 0.00685 & \textbf{0.00550} \\
		 $25\%$ & 85.42\% & 0.00664 & 0.00683 & \textbf{0.00538} \\
		 \hline
	\end{tabular}
\begin{tabular}{c || c | c | c }
	\hline
	Train &  \multicolumn{3}{c}{$\|\A\x_M - \y\|_2$}\\
	$\%$ & Const. & Epoch & NN Pred.\\
	\hline
	$75\%$ & \textbf{0.25087} & 0.26671 & 0.25121 \\
	$50\%$ & 0.25247 & 0.26792 & \textbf{0.24717} \\
	$25\%$ & 0.24885 & 0.26469 & \textbf{0.24323} \\
	\hline
\end{tabular}
\end{center}
	\caption{Average test set validation accuracies of trained neural network (averaged over 10 trials), average resulting recovery error $\|\x_M - \x\|_2$ and residual error $\|\A\x_M - \y\|_2$ on test set for constant learning rate, epoch-based learning rate, and the neural network (NN) predicted learning rates on computerized tomography linear systems.}
	\label{table:CTortho}
\end{table} 

\begin{figure}[t!]
\centering
	\includegraphics[width=0.7\textwidth]{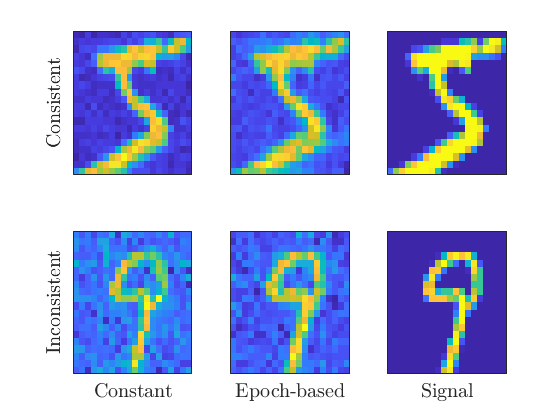}
	\caption{Recovered signals on consistent system using 5000 SGD iterations with constant learning rate (top left; recovery error 0.00278), epoch-based learning rate (top middle; recovery error 0.00555), and original signal (top right).  Recovered signals on inconsistent system using 5000 SGD iterations with constant learning rate (bottom left; recovery error 0.01117), epoch-based learning rate (top middle; recovery error 0.00899), and original signal (bottom right). The neural network correctly predicts the optimal learning rates on these systems.}
	\label{fig:MNISTrecovery}
\end{figure}

In order to visualize the potential improvement offered by using the trained neural net to select optimal step sizes for each tomography system, we plot in Figure \ref{fig:MNISTrecovery} a recovered image using 5000 SGD iterations with each learning rate schedule, and the original image.  The neural network predicts the correct optimal learning rate schedule on these systems.

\begin{figure}[t!]
	\begin{center}
	\includegraphics[width=0.5\textwidth]{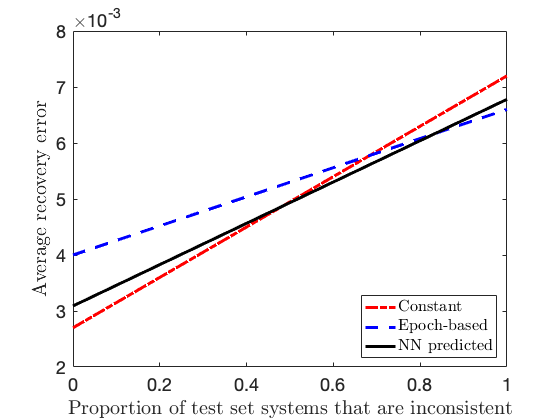}
\end{center}
\caption{The average recovery error on the test set versus the proportion of the test set systems that are inconsistent when using only the constant learning rate, only the epoch-based learning rate, or the 80\% accurate neural network predicted learning rates.}
\label{fig:recoveryerroranalysis}
\end{figure}

These numerical experiments show that a simple neural network trained with proper training data can predict learning rates which improve the recovery error of SGD on a set of given systems. We emphasize that this approach is promising for data sets in which knowledge of the data (e.g., consistency of linear systems, approximate Lipschitz parameters, etc.) is limited.  Depending upon the makeup of the given data set and the choice of learning rate schedules, choosing to use a single schedule on all data sets may be the optimal choice (in average recovery error), but this approach is useful if you do not have much knowledge about the data set.  We illustrate this with a toy situation plotted in Figure \ref{fig:recoveryerroranalysis}.  We plot the average recovery error versus the proportion of the test set systems that are inconsistent.  For this visualization, we use the recovery errors from the experiment in Figure \ref{fig:MNISTrecovery} to approximate the average recovery errors for each learning rate schedule on each set of systems.  For the neural network predicted average recovery error, we assume that the neural network predictions are 80\% accurate on both the inconsistent systems and the consistent systems.  In this toy example, we see that the neural network predictions outperform the other learning rate schedules when the proportion of inconsistent systems in the test set is between approximately 30\% and 80\%.  However, we additionally note that the neural network predicted learning rates never result in a significantly worse average recovery error than the optimal. Thus, if you know very little about your data set (e.g., how many systems are inconsistent) then our framework offers an efficient method to decrease the resulting average recovery error over all data.

\section{Conclusion}
We have presented a machine learning framework for algorithm selection or parameter tuning that is applicable in all areas of computational mathematics.
We showcased its broad potential by applying it to compressed sensing and stochastic gradient descent. 
As long as we have a choice of algorithms, or parameter values that determine the behavior of an algorithm, the same process of training
a neural network can be used to obtain automatic recommendations. This presents the possibility that in the future, software will integrate
some way to collect data in order to improve itself. Futuristic code will adjust its own parameters based on historic experience of executions of prior instances.
We predict this will be useful not only in self-improvement in machine learning (e.g., as stochastic gradient descent improves itself from its
data, it will improve learning models), but it will also be useful in other fields of computational mathematics where a non-expert human 
is at a disadvantage with respect to code that collects lots of data points and self-improves. What are the challenges and limits for this approach? 
They include the right selection of features for training, the amount of data required, and the  type of neural networks used. All of these present interesting mathematical directions.

\section*{Acknowledgements}
This material was also support by the National Science Foundation grant number DMS-1440140 while the authors were in residence at the Mathematical Science Research Institute in Berkeley, California, during the Fall 2017 semester. De Loera was funded by DMS-1522158. Needell was funded by NSF CAREER DMS $\#1348721$ and NSF BIGDATA DMS $\#1740325$. 

\bibliographystyle{abbrv}
\bibliography{bib}

\end{document}